\documentclass[letterpaper]{article} % DO NOT CHANGE THIS
\usepackage[table]{xcolor} 
\usepackage{aaai2026}  % DO NOT CHANGE THIS
\usepackage{times}  % DO NOT CHANGE THIS
\usepackage{helvet}  % DO NOT CHANGE THIS
\usepackage{courier}  % DO NOT CHANGE THIS
\usepackage[hyphens]{url}  % DO NOT CHANGE THIS
\usepackage{graphicx} % DO NOT CHANGE THIS
\usepackage{natbib}  % DO NOT CHANGE THIS AND DO NOT ADD ANY OPTIONS TO IT
\usepackage{caption} % DO NOT CHANGE THIS AND DO NOT ADD ANY OPTIONS TO IT
\usepackage{algorithm}
\usepackage{algorithmic}
\usepackage{pdfpages}
\usepackage{cite}

\usepackage{newfloat}
\usepackage{listings}
\DeclareCaptionStyle{ruled}{labelfont=normalfont,labelsep=colon,strut=off} % DO NOT CHANGE THIS
\floatstyle{ruled}
\newfloat{listing}{tb}{lst}{}
\floatname{listing}{Listing}
\usepackage{multirow}
\usepackage{subcaption}
\usepackage{amsmath}
\usepackage{graphicx}

\usepackage{subcaption}
\nocopyright

\title{ELV-Halluc: Benchmarking Semantic Aggregation Hallucinations in Long Video Understanding}
\author{
    Hao Lu\equalcontrib, Jiahao Wang\thanks{Project Lead}\equalcontrib, Yaolun Zhang, Ruohui Wang, Xuanyu Zheng,\\ Yepeng Tang, Dahua Lin and Lewei Lu\thanks{Corresponding author: luotto@sensetime.com}\\
}
\affiliations{
    SenseTime Research\\
}

\usepackage{bibentry}
\begin{document}

\maketitle

\begin{abstract}
Video multimodal large language models (Video-MLLMs) have achieved remarkable progress in video understanding. However, they remain vulnerable to hallucination—producing content inconsistent with or unrelated to video inputs. Previous video hallucination benchmarks primarily focus on short-videos. They attribute hallucinations to factors such as strong language priors, missing frames, or vision-language biases introduced by the visual encoder. While these causes indeed account for most hallucinations in short videos, they still oversimplify the cause of hallucinations. Sometimes, models generate incorrect outputs but with correct frame-level semantics. We refer to this type of hallucination as \textbf{Semantic Aggregation Hallucination (SAH)}, which arises during the process of aggregating frame-level semantics into event-level semantic groups. Given that SAH becomes particularly critical in long videos due to increased semantic complexity across multiple events, it is essential to separate and thoroughly investigate the causes of this type of hallucination. To address the above issues, we introduce \textbf{ELV-Halluc}, the first benchmark dedicated to long-video hallucination, enabling a systematic investigation of SAH. Our experiments confirm the existence of SAH and show that it increases with semantic complexity. Additionally, we find that models are more prone to SAH on rapidly changing semantics. Moreover, we discuss potential approaches to mitigate SAH. We demonstrate that positional encoding strategy contributes to alleviating SAH, and further adopt DPO strategy to enhance the model’s ability to distinguish semantics within and across events. To support this, we curate a dataset of 8K adversarial data pairs and achieve improvements on both ELV-Halluc and Video-MME, including a substantial 27.7\% reduction in SAH ratio. The dataset and evaluation code can be found at https://github.com/hlsv02/ELV-Halluc.
\end{abstract}


\section{Introduction}
\begin{figure}[t!]
    \centering
    \includegraphics[width=1\linewidth]{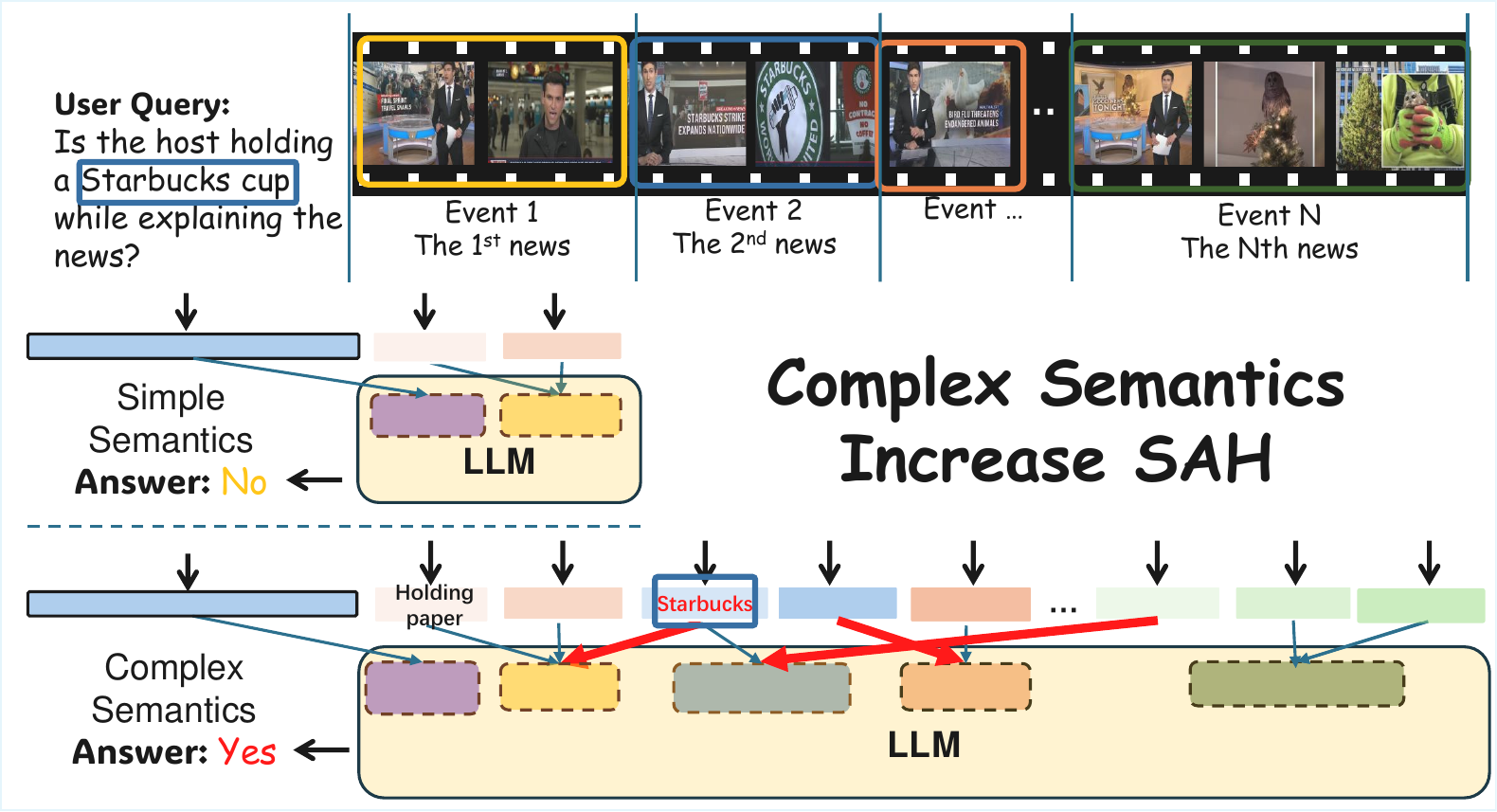}
    \caption{Illustration of how increasing semantic complexity in long-video scenarios amplifies Semantic Aggregation Hallucination (SAH). Red arrows indicate erroneous aggregation into internal semantic groups.}
    \label{fig:1}
\end{figure}
Video multimodal large models have demonstrated strong capabilities in visual understanding\cite{bai2025qwen2.5vl,comanici2025gemini2.5,zhu2025internvl3}. However, a serious challenge still remains—the hallucination, where models generate content that is inconsistent with or even fabricated beyond the video content, thereby impacting the reliability of the models in practical applications. Many works\cite{rawal2025argus,zhang2024eventhallusion,li2025vidhalluc,kong2025mhbench,wang2024videohallucer} have attempted to measure hallucination in video MLLMs, but they primarily focus on short videos ranging from a few seconds to tens of seconds, leaving hallucination issues in long-video contexts largely unexplored. They attribute hallucinations in Video-MLLMs to factors such as vision–language misalignment, poor frame quality, or suboptimal frame sampling strategies, which cause the model to rely on incomplete or inaccurate visual evidence. Alternatively, model may correctly perceive visual semantics but over-rely on strong language priors, disregarding visual input and producing incorrect content. 

While above causes indeed covers large proportion of hallucinations, another cause has been overlooked in prior short video hallucination benchmarks: cases where the model correctly perceives and outputs accurate frame-level semantics but still produces incorrect content by misattributing semantics across event. For example, in Figure \ref{fig:1}, the model attributes ``Starbucks" to the first event, where the host is holding ``some paper" while explaining the news. However, the mention of ``Starbucks" actually corresponds to a later event in the video. In this case, while the perception of frame-level visual semantics is correct, the error arises from misaggregating information across temporal segments—incorrectly linking visual cues from one event to concepts from another. We refer to this phenomenon as Semantic Aggregation Hallucination (SAH).

In short-video scenarios, the impact of SAH is limited because frame-level semantics usually map directly to a single, self-contained event. As a result, logically consistent event semantics—especially
those aligned with language priors—relatively rare (e.g.,``person", ``horse", ``riding"; it is unlikely for the model to hallucinate ``horse riding a person,"). In contrast, long videos often contain multiple temporally extended, yet semantically coherent events, increasing the risk of misattributing concepts across events. As illustrated in Figure \ref{fig:1}, this richer temporal structure amplifies the likelihood of SAH, where the model confuses when an event occurs, even if all the visual elements are correctly perceived.

To address the aforementioned limitations, we introduce ELV-Halluc, the first long video hallucination benchmark. As SAH is particularly prominent and challenging in long videos yet remains underexplored, ELV-Halluc is designed for studying SAH by quantifying semantic complexity through event-based videos and categorizing hallucination aspects based on semantic granularity, including visual details, action, object, and declarative content. To facilitate focused investigation, it adopts an adversarial triplet question pair design: (1) Ground Truth Question paired with In-Video Hallucinated Question, and (2) Ground Truth Question paired with Out-of-Video Hallucinated Question. We use the accuracy gap between in-video and out-of-video hallucinated question pairs to quantify a model’s sensitivity to semantic misalignment across events—a key aspect of SAH. Furthermore, we define the \textbf{SAH Ratio} as the proportion of such cases among all hallucinations, enabling systematic and interpretable analysis.

We conducted extensive experiments on \textbf{ELV-Halluc}, covering 14 open-source MLLMs and 2 closed-source models. Our findings confirm the existence of SAH and reveal that it does not necessarily correlate with overall hallucination rates. Notably, SAH becomes more severe as semantic complexity increases—such as with more events or denser frame sampling—and is more likely to occur in fine-grained, rapidly changing aspects (e.g., visual details rather than declarative content). Since SAH arises from incorrect aggregation of frame-level semantics across events, we show that strengthening the mapping between frames and events—such as through improved positional encodings—can help reduce its occurrence. We further adopt DPO\cite{rafailov2023dpo} strategy that explicitly discourages the model’s preference for hallucinated semantics. Our contributions are summarized as follows:

\begin{itemize}
\item We introduce ELV-Halluc, the first long-video benchmark designed specifically to evaluate SAH.
\item We conduct extensive experiments demonstrating that SAH positively correlates with semantic complexity and semantic variation rate(e.g., more SAH on visual details than declarative content). This relationship causes SAH to sometimes exhibit trends opposite to overall hallucination levels (e.g., when more frames are sampled). 
\item We validate the effectiveness of multimodal positional encoding in mitigating SAH and further adopt DPO strategy to reduce SAH. By curating 8K QA pairs with and without hallucinations, we achieve a maximum 27.7\% reduction in SAH ratio while also improving overall performance (+0.9\% on VideoMME).
\end{itemize}

%Specifically, Video-MLLMs exhibit increased SAH as semantic complexity grows—such as with a higher number of events or denser frame sampling—while denser sampling. Additionally, models are more prone to SAH on rapidly changing and fine-grained semantics(e.g., more SAH on visual details than declarative contents), whereas stronger language models help mitigate SAH. Since SAH originates from the model’s internal confusion when aggregating frame-level semantics across events, we argue that establishing a stronger and more explicit mapping from frames to events can help mitigate SAH. We validate that positional encoding strategies contribute to reducing SAH and propose a DPO-based optimization approach to decrease the model’s preference for hallucinated semantics.

\section{Related Works}
\subsection{Video Understanding Benchmarks}
Video Understanding benchmarks such as Video-MME\cite{fu2025videomme} and MVBench\cite{li2024mvbench} aim to provide a comprehensive evaluation of video understanding capabilities, covering multiple video lengths and diverse aspects of comprehension. Some benchmarks, however, focus on specific abilities of video models; for example, ETBench\cite{liu2024etbench} emphasizes temporal localization and time-awareness, while Video-Holmes\cite{cheng2025videoholmes} evaluates complex reasoning capabilities through QA pairs requiring strong reasoning skills.

In long video contexts, LVBench\cite{wang2024lvbench} evaluates model comprehension for ultra-long videos exceeding one hour, while MLVU\cite{zhou2024mlvu} designs tasks with different requirements—such as holistic understanding, single-detail comprehension, and multi-detail reasoning—to assess long-video capabilities. Similarly, EgoSchema\cite{mangalam2023egoschema} emphasizes evaluating model performance in egocentric video scenarios. However, hallucination—an important and relatively independent aspect of model reliability—remains largely underexplored in these general-purpose video understanding benchmarks.

\subsection{Hallucination Evaluation in Video-MLLMs}
Several prior efforts have aimed to construct hallucination-specific benchmarks. VideoHallucer\cite{wang2024videohallucer} categorizes hallucinations into two types: intrinsic, where the model outputs content inconsistent with the original video, and extrinsic, where correctness cannot be determined solely from the video. EventHallusion\cite{zhang2024eventhallusion} further identifies two main sources of hallucination in Video-MLLMs: language priors and vision-language bias, and investigates these through QA designs involving rare events and misleading contexts. VidHalluc\cite{li2025vidhalluc} focuses on evaluating hallucinations in dynamic video segments and argues that the inductive bias inherent in visual encoders makes hallucinations more likely when processing semantically similar videos. ARGUS\cite{rawal2025argus}, on the other hand, emphasizes hallucination evaluation in open-ended video captioning tasks.

Nevertheless, these existing benchmarks share two major limitations:
(1) They primarily target short videos with relatively simple semantics.
(2) They lack explicit discussion of Semantic Aggregation Hallucination (SAH), a critical challenge in long-video understanding.

\begin{figure*}[h!]
    \centering
    \includegraphics[width=1\linewidth]{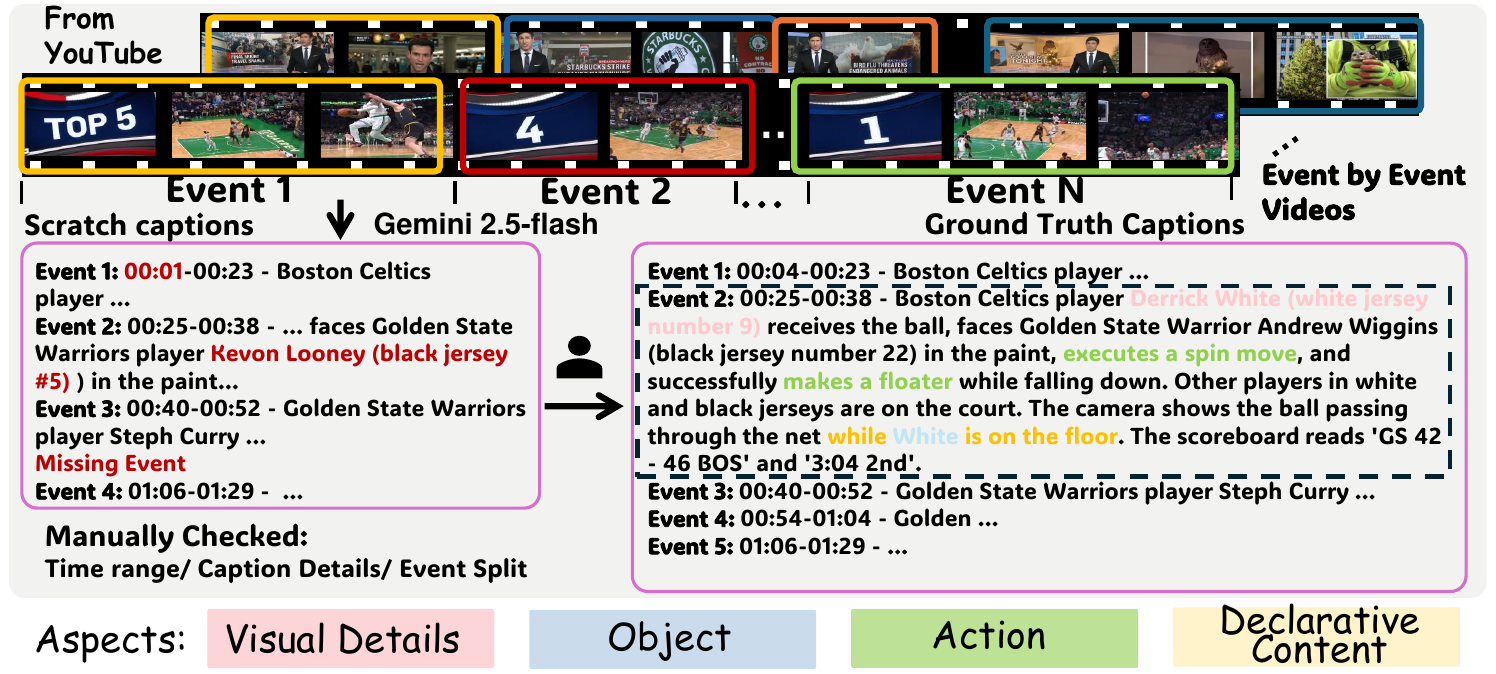}
    \caption{Overview of the data construction pipeline. Scratch captions are first generated using Gemini 2.5 Flash, followed by manual verification to obtain ground-truth captions. Different colors indicate semantic content requiring further modification for hallucination.}
    \label{fig:pipeline}
\end{figure*}
\section{ELV-Halluc}

To address above issues, we propose ELV-Halluc, a Event based Long Video Hallucination benchmark and conduct in-depth analysis on SAH. Table \ref{tab:stat} shows statistical comparison with other video hallucination benchmarks. 
\begin{table}[htbp]
\resizebox{\columnwidth}{!}{%
\centering
\begin{tabular}{cccc}
\hline
\textbf{} & \textbf{Videos} & \textbf{QAs per Video} & \textbf{Avg Video Length (s)} \\
\hline
Videohallucer & 948 & 1.89  & 85.6 \\
EventHallusion & 397    & 1.77     & 11.2 \\
VidHalluc      & 5002 & 1.86  & 24.7 \\
ARGUS          & 500      &   19      & 26.3 \\
\hline
Ours           & 200 & 24  & 672.4 \\
\hline
\end{tabular}%
}
\caption{Statistical comparison with previous video hallucination benchmarks}
\label{tab:stat}
\end{table}
\subsection{Event by event Video collection}
Our benchmark is composed of \textbf{Event-by-Event Videos}. We define event-by-event Videos as videos that consist of multiple clearly separated events sharing the same overall topic. (e.g., a news broadcast with multiple news items).

Event-by-event Videos offer several advantages for establishing a long-video hallucination benchmark:
\begin{itemize}
    \item Videos with clearly separated events can reduce captioning difficulty by isolating semantic units.

    \item Increase the likelihood of inducing SAH, as the semantics of event by event videos can be reorganized into multiple plausible yet incorrect descriptions.
    \item In event by event videos, the number of events can serve as an intuitive indicator of semantic complexity.
\end{itemize}

Finally, we manually collected 500 videos from YouTube. We removed overlapping samples with datasets such as YouCook2 to prevent potential data leakage.
\begin{figure*}[t!]
    \centering
    \includegraphics[width=1\linewidth]{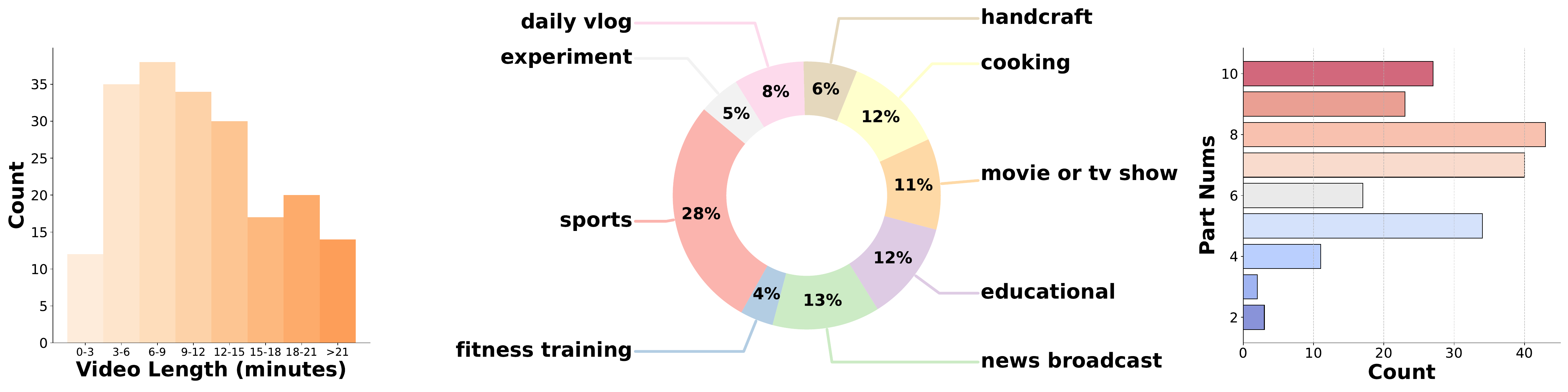}
    \caption{Dataset statistics: duration distribution (left), topic distribution (middle), and number of events per video (right).}
    \label{fig:stats}
\end{figure*}
\subsection{Semi-automated Caption Pipeline}
As shown in Figure \ref{fig:pipeline}, we adopt a three stage semi-automated caption pipeline to reduce human effort while ensuring annotation quality.

\subsubsection{Video Quality Recheck}
To reduce annotator disagreement on the Event-by-Event concept, we conducted a quality recheck. Annotators retained videos with 2–10 clearly distinguishable events and provided a keyword summarizing each video’s core event (e.g., scoring moments in basketball, news coverage in broadcasts). A total of 348 videos were retained, each reviewed by at least two annotators.

%We perform a quality recheck step to minimize annotator disagreement regarding the Event-by-Event concept. Annotators were asked to keep videos which the number of events is within 2–10, and event boundaries are clearly distinguishable.
%In addition, annotators were asked to provide a keyword discribing the event for each video (e.g., scoring moments for basketball games, news event for news videos).
%In the end, we retained 347 videos and each video has been checked by at least 2 distinct annotators.

\subsubsection{Automated Caption Generation with Gemini}
We used Gemini-2.5 Flash to generate initial captions. Gemini was prompted to segment videos based on annotated keywords, exclude transitional or non-essential parts, and produce detailed captions for each identified event.

% For these videos, we employed Gemini-2.5 Flash to generate sratch captions. Specifically, Gemini was instructed to:
% Segment the video into event intervals based on the previous annotated keywords;
% Exclude transitional segments and introduction/conclusion parts;
% Produce detailed captions for each identified event.

\subsubsection{Human Verification and Refinement}
Annotators are asked to revise the gemini generated captions through the following steps:
1. Correcting inaccurate time ranges;
2. Fixing factual errors in the captions;
3. Removing redundant segments (e.g., introductions, summaries, transition parts);
4. Adding missing events and manually annotating captions.

Following this semi-automated process, we obtained 348 high-quality Event-by-Event videos with human-refined ground truth captions, ensuring accuracy while substantially reducing manual annotation costs.

\subsection{Hallucinate QAs curation}
We design adversarial question pairs for better hallucination evaluating. A model should be able to both chose correct caption and reject hallucinated captions. Based on this principle, we use GPT-4o to modify ground truth captions by introducing hallucination elements. Each modified caption is paired with its ground truth to form a Question pair, and the pair is considered correct only if model answers both questions correctly.

Our modifications specifically target the semantics of four aspects:
\textbf{Visual details}: attributes such as color, shape, size, patterns, spatial relationships, or on-screen text (OCR).
\textbf{Action}: denotes the key activity or motion being performed.
\textbf{Object}: refers to humans or physical objects mentioned in the caption.
\textbf{Declarative content}: descriptive or propositional statements summarizing a situation, asserting an outcome, or conveying a belief or result, rather than concrete actions or events. (e.g., ``Team A is leading," ``The match seems intense,")

\begin{figure}[htbp]
    \centering
    \includegraphics[width=1\linewidth]{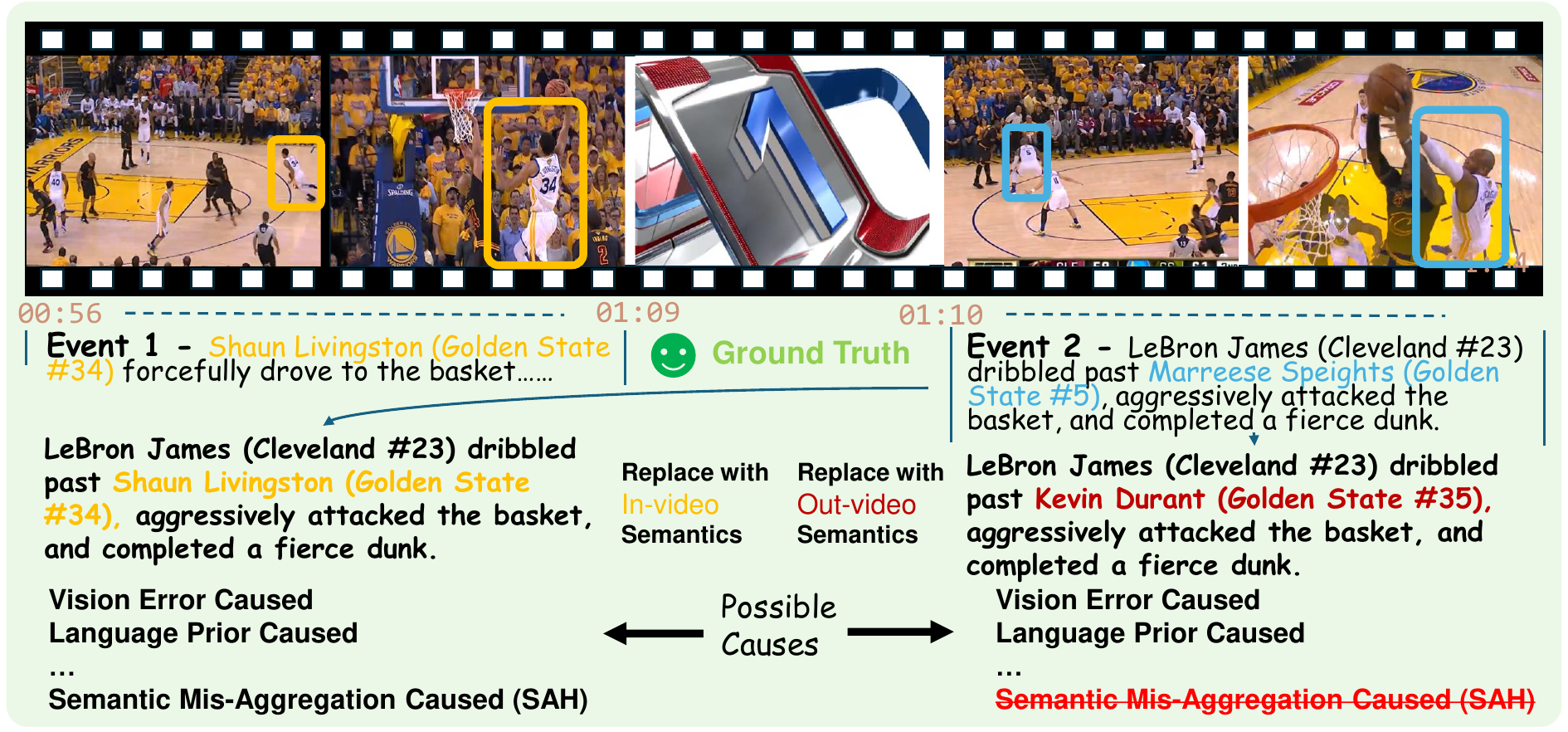}
    \caption{Showcase of in-video and out-video modifications.}
    \label{fig:2}
\end{figure}
GPT-4o is instructed to modify captions by altering one of these aspects.
To further investigate SAH, we design two modification strategies:

\textbf{In-video modification}: GPT replaces an object in the ground truth caption with an object drawn from another event within the same video. 

\textbf{Out-video modification}: GPT replaces an object in the ground truth caption with a fabricated object that does not appear in any captions from the video.

Captions after modification must remain plausible and reasonable, such that correctness cannot be judged without watching the video.
If a model is misled by an in-video hallucinated caption, all hallucination types could be responsible causes. In contrast, if the model is misled by an out-video hallucinated caption, SAH won't serve as  possible cause, since the hallucinated content does not exist in the video. Therefore, \textbf{subtracting the out-video mislead rate from the in-video mislead rate approximates the contribution of Semantic Aggregation Hallucination.}As shown in Figure \ref{fig:2}, We use an object as an example to demonstrate the In-video and Out-video modifications.

After applying these modifications, we obtain 20072 hallucinated captions across 348 videos.

\subsection{Hallucinated Caption Quality Check}
We used GPT-4o to automatically recheck all modified captions, ensuring that in-video captions introduce the desired aspect change present in other events’ ground truths, while out-video captions introduce changes absent from all ground truths. Captions meeting above criteria were retained, yielding 348 Event-by-Event videos and 8,630 hallucinated caption pairs.
\begin{figure}[htbp]
    \centering
    \includegraphics[width=1\linewidth]{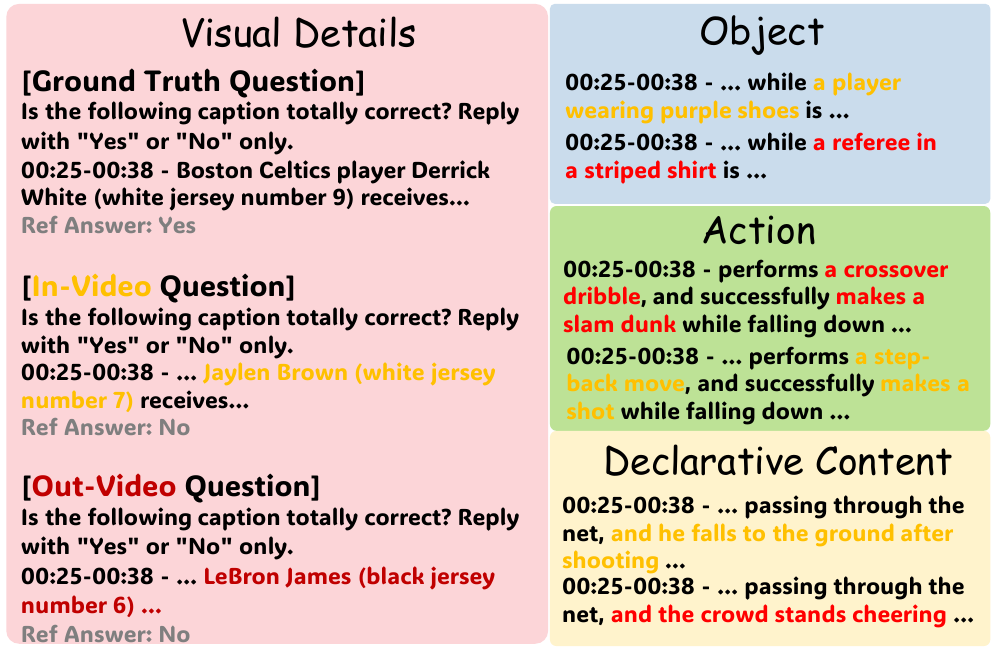}
    \caption{Examples of ELV-Halluc question–answer pairs across different semantic aspects. The left part illustrates the complete QA pair format in ELV-Halluc.}
    \label{fig:finalshowcase}
\end{figure}

\subsection{Final Benchmark and Evaluation Metrics}
We select 200 videos from the original set of 348, leaving the remaining 148 videos as the training set for DPO. For each selected video, we choose 24 captions to construct binary QA pairs by appending the question prefix:
\textit{``Is the following caption totally correct? Reply with ‘Yes’ or ‘No’ only.''}

These QA pairs cover the aforementioned four aspects: visual details, objects, actions, and declarative content. Each aspect includes 6 questions, formed from 2 triplets drawn from different events within the same video. Each triplet consists of three captions: ground truth, in-video hallucinated, and out-of-video hallucinated.
Examples of the final QA pairs are shown in Figure \ref{fig:finalshowcase}.

We form adversarial QA pairs by combining a ground-truth question with a hallucinated question, resulting in two pairs per triplet:
(GT, In-Video Hallucination)
(GT, Out-of-Video Hallucination)
A pair is considered correct only if the model predicts ``Yes” for the ground-truth question and ``No” for the hallucinated question.
Overall, the benchmark contains 4,800 binary QA pairs, which can be further grouped into 3,200 adversarial QA pairs.
Figure \ref{fig:stats} presents detailed statistics of ELV-Halluc, illustrating its diversity in video length, topics, and number of events.
\subsubsection{Accuracy}
We use Accuracy to evaluate the overall hallucination level of models in long-video scenarios. Specifically, we report the following metrics:
In-Video Accuracy: Accuracy on QA pairs containing in-video hallucinations.
Out-Video Accuracy: Accuracy on QA pairs containing out-of-video hallucinations.

\subsubsection{SAH Ratio}
We further propose the \textbf{SAH Ratio} to quantify the proportion of Semantic Aggregation Hallucination (SAH) among all hallucination errors. If a model achieves high accuracy on out-of-video hallucinations but significantly lower accuracy on in-video ones, it indicates difficulty in resolving semantic misalignment across events—the hallmark of SAH. Therefore, the accuracy gap reflects how prone the model is to confusing correct frame-level content with incorrect event-level attribution, making it a suitable proxy for SAH severity. Therefore, we use the SAH Ratio instead of the absolute difference between Out-Video and In-Video accuracy. This approach enables a more precise measurement of the relative severity of SAH while minimizing the influence of the model’s absolute performance level. Consequently, it facilitates targeted solutions specifically addressing SAH. The metric is computed as follows:

\[
\text{SAH Ratio} = \frac{\text{OutAcc} - \text{InAcc}}{1 - \text{InAcc}}
\]

where \textbf{OutAcc} and \textbf{InAcc} denote the accuracy on out-of-video and in-video hallucination pairs, respectively. 
%This metric allows us to measure the relative contribution of SAH to overall hallucination errors.

\section{Experiments and discussions}
\begin{table*}[ht!]
\centering
\renewcommand{\arraystretch}{1.3}
\resizebox{\textwidth}{!}{%
\begin{tabular}{ccccccccccccccccc}
\hline
\multirow{2}{*}{\textbf{Models}}  &  \multirow{2}{*}{\textbf{LLM size}} &
\multicolumn{3}{c}{\textbf{Visual Details}} &
\multicolumn{3}{c}{\textbf{Object}} &
\multicolumn{3}{c}{\textbf{Action}} &
\multicolumn{3}{c}{\textbf{Declarative Content}} &\multirow{2}{*}{\textbf{Avg Acc↑}}
& \multirow{2}{*}{\textbf{Avg Diff.↓}}&\multirow{2}{*}{\textbf{SAH Ratio↓}}\\
& &In. & Out. & Diff. &
  In. & Out. & Diff. &
  In. & Out. & Diff. &
  In. & Out. & Diff. &
   &  & \\
\hline
\multicolumn{16}{c}{\textbf{Open Source Models}} \\
\hline
InternVL3-1B  & 0.5B & 8 & 11 & 3 & 8.7& 11 & 2.3 & 8.7 & 12.5 & 3.8 & 11.3 & 8.3  & -3 & 9.9  & 1.5  & 1.6 \\
InternVL3-2B  & 1.5B & 7& 15.5 & 8.5 & 8.7  & 17.2 & 8.5  & 7.2  & 10.5 & 3.3 & 10 & 13 & 3 & 11.1 & 5.8  & 6.3 \\
SmolVLM-2.2B & 1.7B&0 & 0& 0& 3& 5& 2& 0& 0& 0& 0& 0& 0& 1& 0.5&0.5\\
Qwen2.5VL-3B &3B& 2.2& 10.5& 8.3& 7.7& 13.8& 6.1& 5& 8& 3& 6& 6& 0& 7.4& 4.3&4.5\\
LLaVA-Video-7B &7B& 3.7&3.7&0&4.5&2.5&-2&3.7&3.2&-0.5&4&4&0&3.6&-0.6&-0.6\\
Video-chatgpt-7B &7B& 2&2.5&0.5&2.5&1.7&-0.7&1.2&1.2&0&2.2&3.2&1&2.0&0.2&0.1\\
LLaVA-OV-7b & 7B&8& 13.2&5.2&9.5&13.7&4.2&8.7&10.7&2&7.7&7.5&-0.2&9.9&2.8&3.0\\
Qwen2.5VL-7b &7B& 10.2 & 26& 15.8& 17.5& 30.7& 13.2& 13& 20.7& 7.7& 16.8& 10.5& -6.3& 18.1&7.6 &8.8\\
InternVL3-8B &7B& 12.5 & 19.5 & 7.0 & 14.5 & 19.5 & 5.0 & 13.5 & 20.5 & 7.0 & 12.8 & 17.7 & 4.9 & 16.3 & 5.9 & 6.8 \\
InternVL3-14B &14B& 17.5 & 24.5 & 7.0 & 22.8 & 24.5 & 1.7 & 16.3 & 17.7 & 1.4 & 15.2 & 15.5 & 0.3 & 19.2 & 2.6 & 3.1 \\
Qwen2.5VL-32B &32B& 16.5 & 24.5 & 8.0 & 21.7 & 24.5 & 2.8 & 17.2 & 15.0 & -2.2 & 15.2 & 7.2 & -8 & 17.7 & 0.1 & 0.2 \\
InternVL3-38B &32B& 25.3 & 29   & 3.7  & 24.2 & 28   & 3.8  & 24   & 30   & 6    & 24.5 & 24.2 & -0.3  & 26.1 & 3.3  & 4.3 \\
Qwen2.5VL-72B &72B& 24 & 35.5 & 11.5 & 35.7 & 41.5 & 5.8 & 27.8 & 32.3 & 4.5 & 32.3 & 27 & -5.3 & 32.0 & 4.1 & 5.8 \\
InternVL3-78B &72B& 25 & 31.2 & 6.2 & 32 & 36.5 & 4.5 & 28.5 & 31.2 & 2.7 & 24.2 & 26.5 & 2.3 & 29.3 & 3.9 & 5.4 \\

\hline
\multicolumn{16}{c}{\textbf{Closed Source Models}} \\
\hline
GPT-4o &/& 7.7&8.3&0.6&8&8.7&0.7&8.7&10.2&1.5&8.5&9.5&1&8.7&0.9&1.0\\
\rowcolor{gray!20} 
Gemini2.5-Flash &/& 47&58&11&56.5&58.8&2.3&50.5&53.2&2.7&48.7&52&3.3&53.1&4.8&9.8 \\
\hline
\end{tabular}
}
\caption{Main results on ELV-Halluc. Diff. denotes the gap between in-video and out-video accuracy. Note that the semi-automatic annotation pipeline uses Gemini for initial captioning, which may introduce bias; therefore, metrics for Gemini-2.5 Flash should not be directly compared with other models. All accuracies are shown as percentages.}
\label{tab:main_results}
\end{table*}
We evaluate 14 open-source models ranging from 1B to 78B parameters, along with two closed-source MLLMs on ELV-Halluc. As shown in Table \ref{tab:main_results}, the results demonstrate that current SOTA LLMs continue to face significant challenges with hallucination in long-video understanding(Note: The results of Gemini-2.5 Flash should not be directly compared with other models, as it was used to generate the initial captions, which may introduce bias). The results also validate the existance of SAH, as most MLLMs exhibit substantially lower accuracy on in-video hallucinated captions than on out-of-video hallucinated captions. Among the open-source models, Qwen2.5-VL-32B achieves the lowest SAH Ratio, at only 0.2\%.

\subsubsection{SAH increase as the semantic complexity rises.}
\begin{figure}[h!]
    \centering
    \begin{subfigure}[b]{0.48\linewidth}
        \centering
        \includegraphics[width=\linewidth]{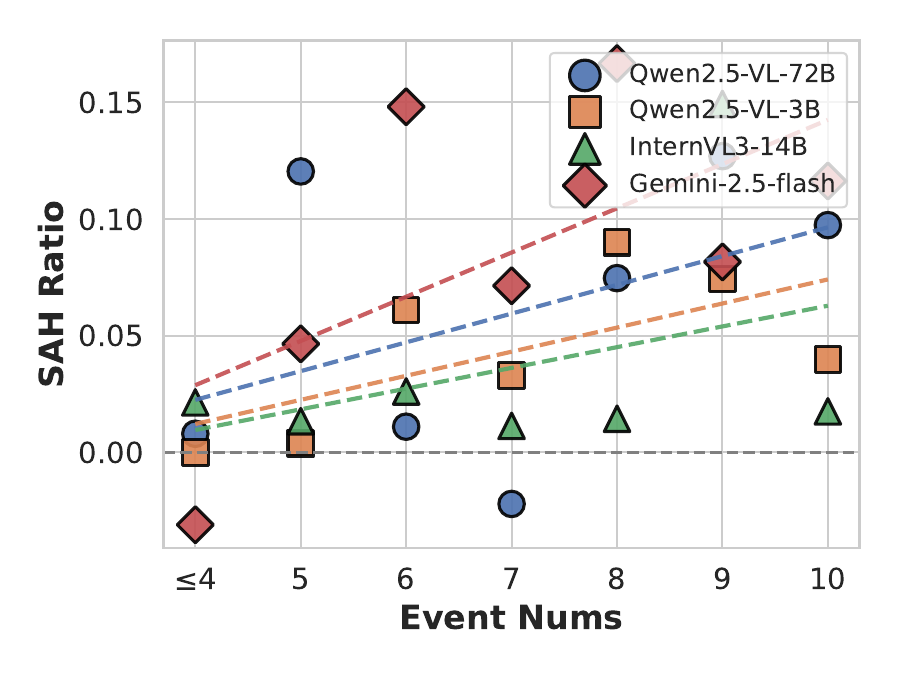}
        \caption{SAH Ratio vs. Event Count}
        \label{fig:sah-event}
    \end{subfigure}
    \hfill
    \begin{subfigure}[b]{0.48\linewidth}
        \centering
        \includegraphics[width=\linewidth]{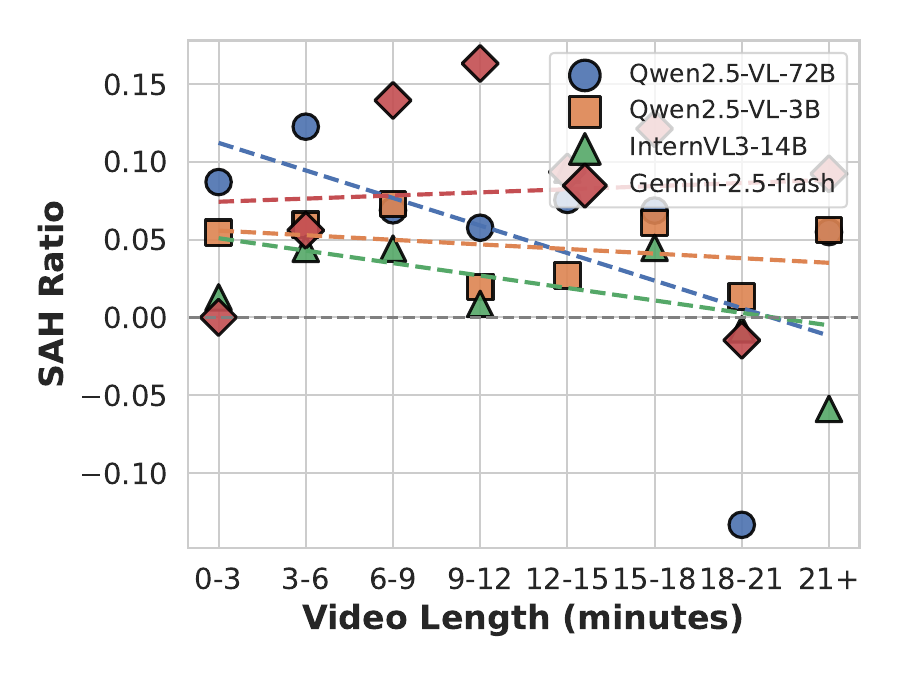}
        \caption{SAH Ratio vs. Video Length}
        \label{fig:sah-time}
    \end{subfigure}
    \caption{Relationship between SAH and video properties: (a) SAH Ratio correlation with the number of events; (b) SAH Ratio correlation with video duration.}
    \label{fig:sah-analysis}
\end{figure}

In Figure \ref{fig:sah-analysis}, we observe that Qwen2.5-VL-3B, Qwen2.5-VL-72B, InternVL3-8B, and Gemini2.5-Flash (representing different model sizes and families) exhibit a positive correlation between the SAH ratio and the number of video events, as a larger number of events typically introduces more complex semantics. Meanwhile, we find that the SAH ratio does not show any consistent relationship with video length, since in our event-by-event dataset, video length does not necessarily correspond to a higher number of events.

\subsubsection{SAH Occurs More Frequently at Rapidly Changing Semantic}
We found that model tend to exhibit SAH more frequently on semantics that change rapidly over time. During modification, we define four aspects—visual details, action, object, and declarative content—each representing a different level of semantic granularity. Among them, visual details vary the fastest, followed by actions, as objects often engage in multiple actions over time. Objects themselves change less frequently, while declarative content represents higher-level semantics that evolve the slowest.

\begin{figure}[htbp]
    \centering
    \includegraphics[width=0.6\linewidth]{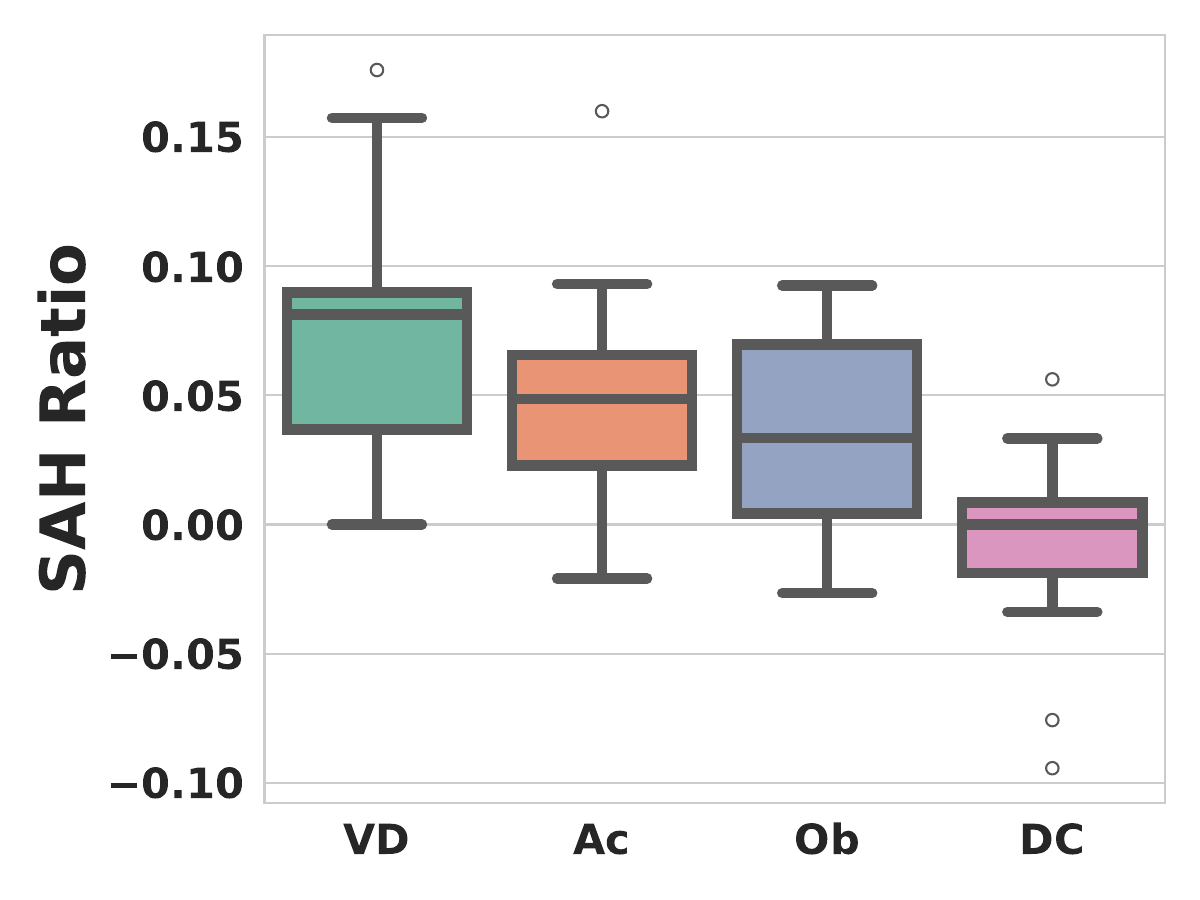}
    \caption{SAH Ratio across different semantic aspects. VD, Ac, Ob, and DC represent Visual Details, Action, Object, and Declarative Content, respectively.}
    \label{fig:boxplot}
\end{figure}

Based on the results of 14 open-source models, we plot the SAH ratio for these four aspects (see Figure \ref{fig:boxplot}). Results reveal that Video-MLLMs exhibit the highest SAH ratio on visual details, followed by actions and objects, and the lowest on declarative content, suggesting that models are more prone to SAH on semantics with higher temporal variability.

\subsubsection{Hallucination under Varying Frame Numbers and Model Sizes.}

\begin{figure}[t!]
  \centering
  \begin{subfigure}[b]{0.48\linewidth}
    \includegraphics[width=\linewidth]{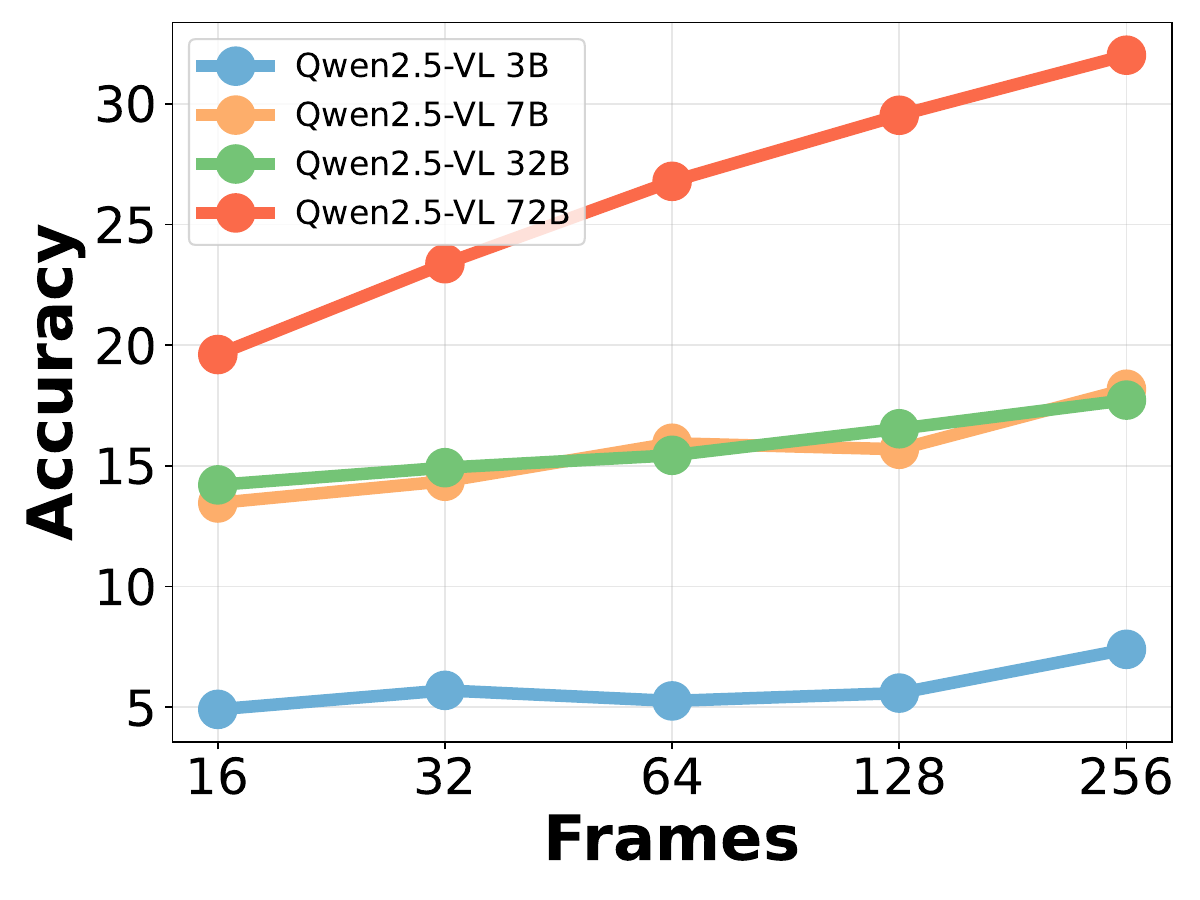}
    \caption{Qwen2.5-VL Acc vs Frame}
    \label{fig:acc_qwen}
  \end{subfigure}
  \hfill
  \begin{subfigure}[b]{0.48\linewidth}
    \includegraphics[width=\linewidth]{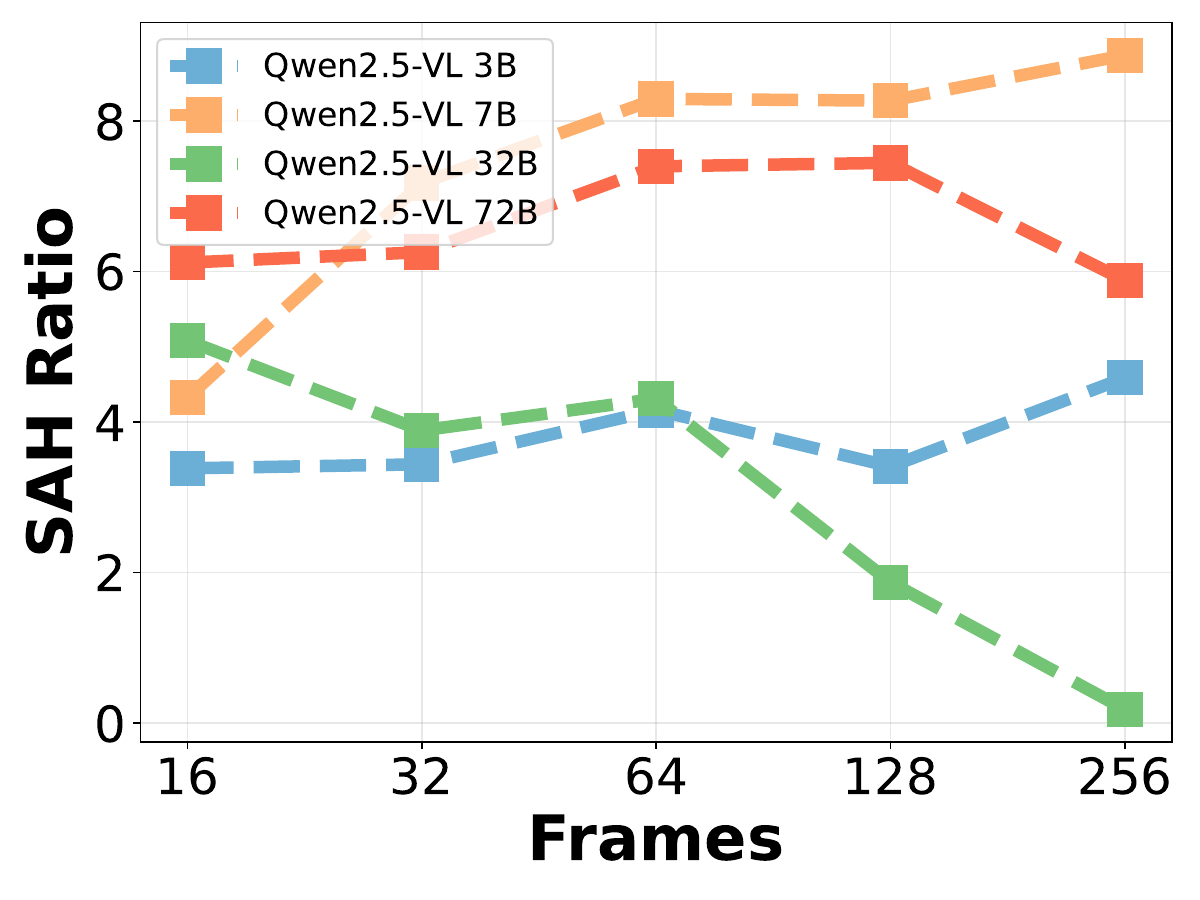}
    \caption{Qwen2.5-VL SAH vs Frame}
    \label{fig:sah_qwen}
  \end{subfigure}
  \begin{subfigure}[b]{0.48\linewidth}
    \includegraphics[width=\linewidth]{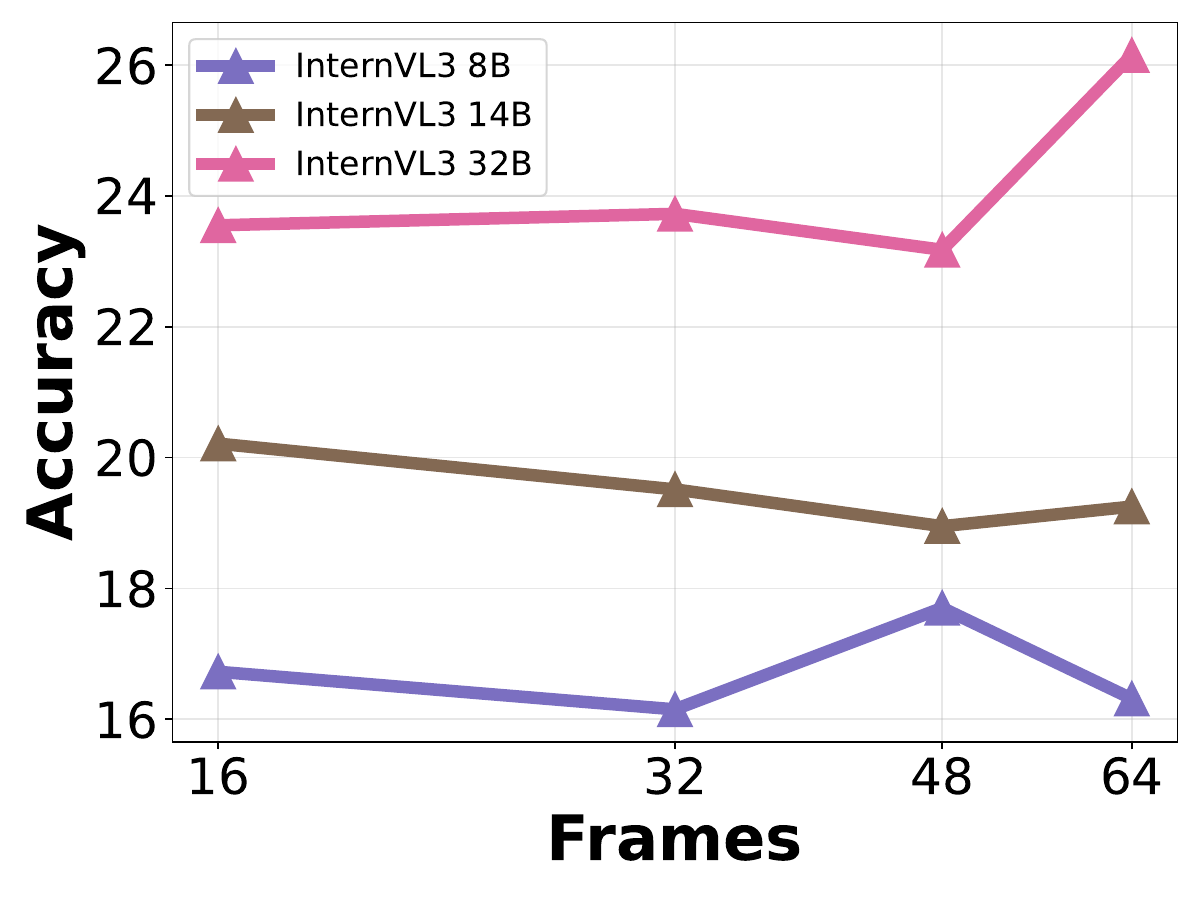}
    \caption{InternVL3 Acc vs Frame}
    \label{fig:acc_intern}
  \end{subfigure}
  \hfill
  \begin{subfigure}[b]{0.48\linewidth}
    \includegraphics[width=\linewidth]{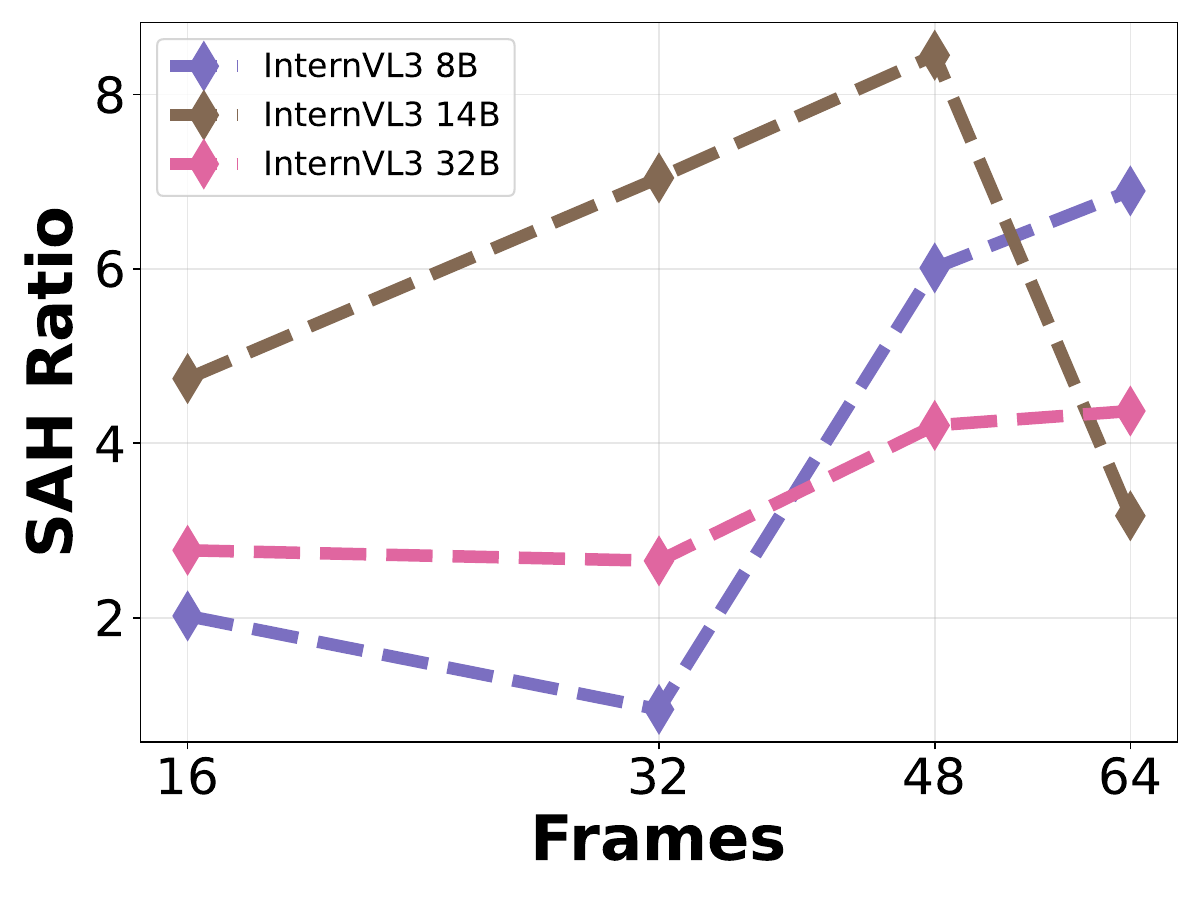}
    \caption{InternVL3 SAH vs Frame}
    \label{fig:sah_intern}
  \end{subfigure}

  \caption{Effect of frame nums and model size on hallucination performance. Each row corresponds to a model family (Qwen2.5-VL and InternVL3), showing (left) overall hallucination accuracy and (right) SAH ratio across varying frame numbers for different model sizes.}
  \label{fig:hallu_frame_performance}
\end{figure}

As shown in Figure \ref{fig:hallu_frame_performance}, We use the Qwen2.5-VL and InternVL3 series models to investigate the relationship between the number of video frames sampled and the occurrence of hallucinations. For Qwen2.5-VL, we disable the dynamic resolution mechanism to ensure that the number of frames serves as the sole varying factor. For accuracy, we observe that for most models, increasing the number of frames generally leads to higher overall hallucination accuracy. For the SAH ratio, most models tend to exhibit higher values as the number of frames increases. We attribute above findings to the fact that an increased number of frames provides the model with richer and more complex semantic information. On one hand, this additional information reduces uncertainty about the overall video content, thereby improving the model’s robustness against overall hallucinations. On the other hand, the added complexity increases the likelihood of semantic mismatches, leading to more SAH.
Interestingly, the Qwen2.5-VL-32B model exhibits a different trend. We hypothesize that this may be due to the reinforcement learning-based post-training of Qwen2.5-VL-32B-Instruct, which likely enhances its ability to aggregate visual semantics effectively.

Furthermore, we observe that larger language models generally achieve higher overall hallucination accuracy, suggesting improved robustness against global hallucinations. However, we do not observe a consistent correlation between model size and the SAH ratio, indicating that increasing model capacity does not necessarily mitigate semantic aggregation hallucinations.

\section{Strategies for Mitigating SAH}
\subsubsection{Position Encoding Decease SAH}

As SAH occur due to errors introduced during the semantic grouping process. We argue that a stronger RoPE mechanism could enhance the model’s ability to bind semantic relationships, thereby reducing errors that occur during grouping. Our experiments validate this hypothesis, as shown in Table \ref{tab:rope_comparison}. Specifically, we evaluate checkpoints based on Qwen2-VL, applying different positional encoding strategies: vanilla RoPE\cite{su2024rope}, TAD-RoPE\cite{gao2024tad-rope}, m-RoPE\cite{Qwen2VL}, and VideoRoPE\cite{wei2025videorope}. The results demonstrate that VideoRoPE achieves the lowest SAH ratio. However, the findings also indicate that stronger RoPE variants do not necessarily mitigate oveall hallucinations.
\begin{table}[ht]
\centering
\resizebox{0.8\columnwidth}{!}{%
\begin{tabular}{lccc}
\hline
\textbf{Method} & \textbf{In. ↑} & \textbf{Out. ↑} & \textbf{SAH Ratio↓}\\
\hline
vanillarope  & 0.94 & \textbf{2.75} & 1.82\\
tad\_rope    & 0.44 & 2.62 & 2.18\\
mrope        & 1.12 & 2.06 &  0.95\\
videorope    & \textbf{1.19} & 2.06 & \textbf{0.88}\\
\hline
\end{tabular}
}
\caption{Different RoPE Strategies on ELV-Halluc. Bold values indicate the best performance. In. and Out. denote the average in-video and out-video accuracy, respectively.}
\label{tab:rope_comparison}
\end{table}
\subsubsection{Mitigate SAH with DPO}
Considering that SAH mainly arise from incorrect grouping of correct video semantics, applying a method to suppress the model's attention to hallucinatory semantics should reduce SAH possibilities. Therefore, we adopt the Direct Preference Optimization\cite{rafailov2023dpo} approach to further optimize the model.

We leverage the remaining 148 videos' ground-truth and hallucinated caption of all events to construct the positive and negative response pairs required for DPO. Specifically, we use the following template to generate data pairs:
``Please provide a detailed caption for this segment during mm:ss - mm:ss.
Chosen: Ground truth caption 
Rejected:Hallucinated caption"

Finally, we create two separate datasets of 4k positive-negative pairs each: one using in-video hallucinated captions and the other using out-of-video hallucinated captions.

We use Qwen2.5-VL-7B as the base model and conduct three training settings:
1. In-video pairs,
2. Out-video pairs,
3. Combining above 2 types of pairs.

\begin{table}[ht]
\centering
\renewcommand{\arraystretch}{1.2}
\resizebox{\columnwidth}{!}{
\begin{tabular}{l|cc|cccc}
\hline
\multirow{2}{*}{\textbf{Model}} & \multicolumn{2}{c|}{\textbf{ELV-Halluc}} & \multicolumn{4}{c}{\textbf{VideoMME}} \\
\cline{2-7}
 & \textbf{Avg Acc↑}  & \textbf{SAH Ratio↓} & \textbf{Short} & \textbf{Medium} & \textbf{Long} & \textbf{Avg↑} \\
\hline
Qwen2.5vl-7B   & 15.9 &  8.3 & 72.7 & 61.7 & 51.3 & 61.9 \\
+ invideo-4k       & 16.2 &  \textbf{6.0} (-27.7\%) & 72.4 & 62.6 & 51.9 & 62.3 \\
+ outvideo-4k       & 16.0 &  8.6 (+3.6\%) &  73.8 & 62.0 & 52.6 & \textbf{62.8} \\
+ together-8k    & \textbf{16.4} &  8.4 (+1.2\%) &  73.4 & 62.1 & 51.8 &  62.4 \\
\hline
\end{tabular}
}
\caption{Performance comparison of base model and DPO variants on ELV-Halluc and VideoMME benchmarks.}
\label{tab:dpo}
\end{table}
As shown in Table \ref{tab:dpo}, applying DPO with in-video pairs yields the most notable improvement, reducing the SAH ratio from 8.3 to 6.0. This indicates that aligning the model’s preference toward correct event semantics within the same video is highly effective for mitigating semantic aggregation hallucinations. Additionally, the overall average accuracy on ELV-Halluc improves by 0.3 points, while general video understanding performance on VideoMME also improves slightly (+0.4), demonstrating that mitigating hallucination does not compromise general capability.

In contrast, DPO with out-of-video pairs reduces overall hallucination slightly but unexpectedly increases the SAH ratio. This suggests that optimizing the model to reject content entirely irrelevant to the video does not effectively improve SAH and may even bias the model toward over-reliance on language priors.

When combining in-video and out-of-video pairs (8k samples together), the model achieves a balanced trade-off: it retains most of the benefits of in-video optimization, while maintaining robustness against out-of-video hallucinations. However, the combined setting does not surpass in-video DPO in reducing SAH.

\begin{figure}[htbp]
    \centering
    \includegraphics[width=1\linewidth]{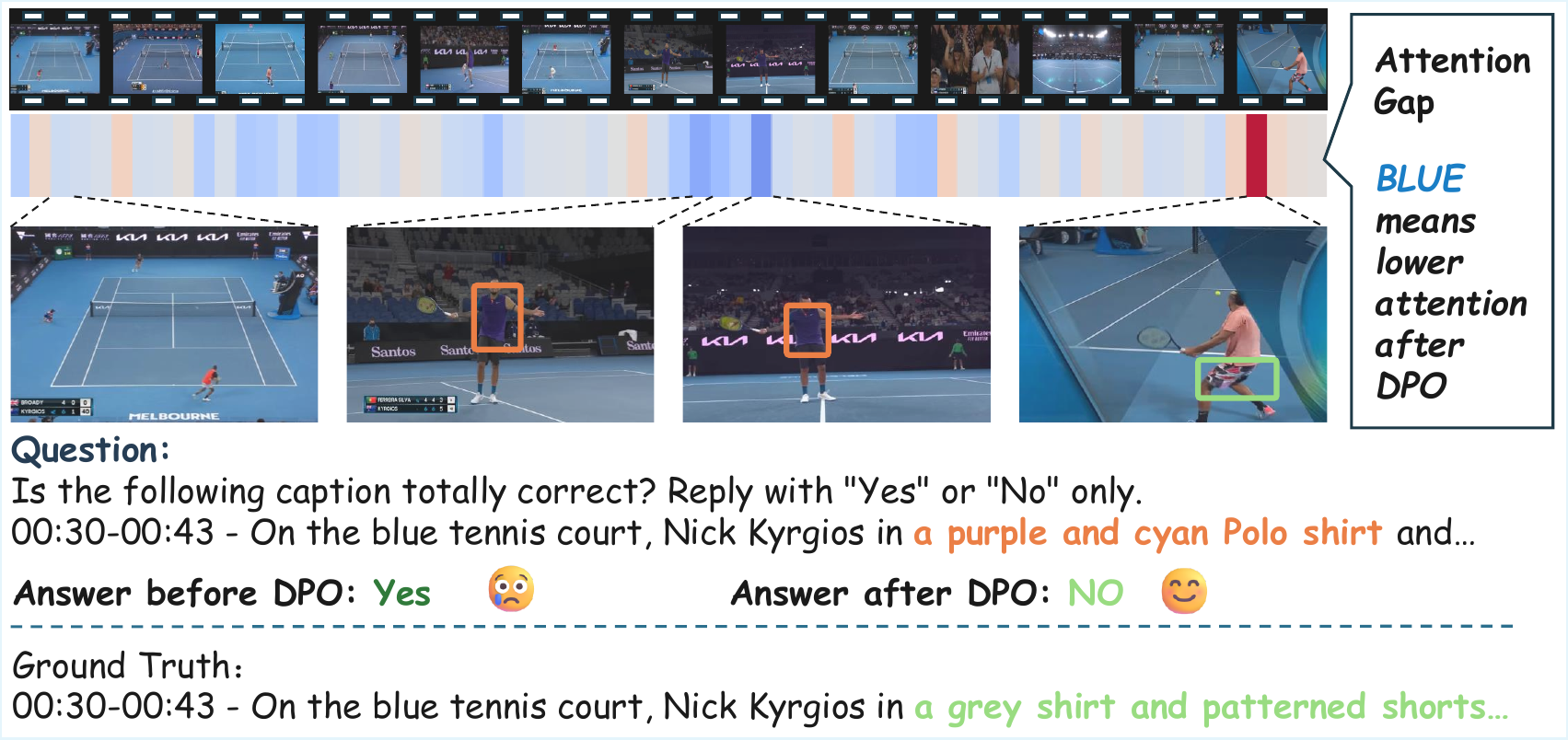}
    \caption{Attention Gap Heatmap after DPO with In-video pairs, BLUE means lower attention after DPO.}
    \label{fig:attn}
\end{figure}
As illustrated in Figure \ref{fig:attn}, attention visualization reveals that after DPO with in-video pairs, the model significantly reduces its focus on incorrect yet semantically plausible regions, shown in blue areas indicating decreased attention weights. This suggests that DPO effectively reshapes the model’s internal preference distribution, promoting stronger grounding of responses in relevant visual semantics. Such alignment at the attention level provides a mechanistic explanation for the observed improvement in SAH mitigation.

\section{Conclusion and Limitations}
In this work, we addressed the underexplored challenge of hallucination in long-video understanding by introducing ELV-Halluc, the first benchmark tailored to evaluate Semantic Aggregation Hallucination (SAH). We identified SAH as a distinct and increasingly prominent error type in semantically complex, multi-event videos. To enable comprehensive evaluation, we benchmarked 14 open-source Video-MLLMs (1B–78B) and two closed-source models (GPT-4o and Gemini 2.5 Flash). Our experiments revealed key SAH patterns and overall hallucination trends. To mitigate SAH, we proposed a DPO-based approach and curated an 8K-pair adversarial dataset, achieving a 27.7\% reduction in SAH and a 0.9\% gain on VideoMME.

% In this work, we tackled the underexplored problem of hallucination in long-video understanding and introduced ELV-Halluc, the first benchmark specifically designed to evaluate Semantic Aggregation Hallucination(SAH). We identified SAH as a critical and distinct type of hallucination that becomes increasingly prominent as semantic complexity grows across multiple events. To establish a comprehensive evaluation, we assessed 14 open-source Video-MLLMs ranging from 1B to 78B parameters, as well as two closed-source models: GPT-4o and Gemini 2.5-Flash. Our extensive experiments revealed overall hallucination trends and provided several insights for SAH. To mitigate this issue, we proposed a DPO-based strategy and curated a 8K-pair adversarial dataset, which effectively reduced SAH by 2.3\% while improving general performance by 0.9\% on VideoMME. 

Despite its contributions, our work has several limitations. First, although our semi-automated pipeline reduces manual effort, Gemini-generated captions may introduce bias, potentially inflating Gemini 2.5 Flash’s performance. Second, while event-based video construction improves control and diversity, it still differs from real-world long videos. Third, the dataset scale is limited due to the high annotation cost. Nonetheless, we believe our benchmark, analysis, and mitigation strategies lay a solid foundation for advancing reliable long-video understanding.

%Despite above contributions, our work also has several limitations:
%1. Although our semi-automated data pipeline reduces manual effort, captions generated by Gemini may introduce bias, potentially leading to an overestimation of Gemini 2.5 Flash’s performance on ELV-Halluc.
%2. While event-by-event video construction offers certain benefits and we have strived to maintain diversity, these videos still exhibit a domain gap compared to real-world long videos.
%3. The overall data scale remains limited due to the high cost of long video annotation.
%Nevertheless, we believe that our benchmark, analysis, and mitigation strategies provide a strong foundation for future research aimed at building more reliable and grounded long-video understanding systems.

\nocite{*}

\bibliography{aaai26}
\clearpage


\section*{Supplementary material}

# ELV-Halluc: Benchmarking Semantic Aggregation Hallucinations in Long Video Understanding

## APPENDIX

- **A: Experiment Setting**
  - A.1: Model Implementation Details
  - A.2: Test Stability Across Multiple Runs
- **B: More ELV-Halluc Cases**
  - B.1: ELV-Halluc QA Examples
  - B.2: ELV-Halluc SAH Cases
- **C: Data Construction Details**

## Experiment Settings

### Model Implementation Details

**Model Checkpoints** For all evaluated open-source models, we adopt the official weights provided on HuggingFace. For the experiments involving RoPE, we directly utilize the Qwen2-VL-based implementation and weights from VideoRoPE. The specific HuggingFace repository paths, along with the versions of the closed-source models, are listed in Table 1.

**Inference Parameters** We provide a detailed description of the parameter configurations used during evaluation. For the Qwen2.5-VL series, we set the maximum video resolution to $128 \times 28 \times 28$ pixels and the maximum number of frames to 256, while all other settings follow the official defaults. For the InternVL3 series, we limit the number of frames to 64. For GPT-4o, we uniformly sample 50 frames and use the following instruction:

*"Here are {nframes} frames sampled from a {video_length}-duration video."*

For all remaining models, we adopt the default generation configurations provided in their respective official checkpoints.

**DPO Experiment Setting** All experiments are conducted on NVIDIA H100 GPUs. During the DPO training process, we train for one epoch with a learning rate of $1 \times 10^{-6}$ and a batch size of 16. We use Qwen2.5-VL-7B as the base model, with the sequence length set to 32,768 during training. All original parameters (e.g., `fps`, `fps max frames`, `video max pixels`) are preserved. During inference, we fix `nframes` to 64 for both **ELV-Halluc** and **VideoMME** benchmarks.

### Test Stability Across Multiple Runs

As shown in Figure 1, we selected four models from different sizes and series for repeated experiments to evaluate the stability of ELV-Halluc across three separate trials. The results demonstrate that ELV-Halluc achieves relatively stable overall accuracy and SAH ratio.

| Model | Checkpoint |
|---|---|
| Qwen2.5-VL-3B | Qwen/Qwen2.5-VL-3B-Instruct |
| Qwen2.5-VL-7B | Qwen/Qwen2.5-VL-7B-Instruct |
| Qwen2.5-VL-32B | Qwen/Qwen2.5-VL-32B-Instruct |
| Qwen2.5-VL-72B | Qwen/Qwen2.5-VL-72B-Instruct |
| InternVL3-1B | OpenGVLab/InternVL3-1B |
| InternVL3-2B | OpenGVLab/InternVL3-2B |
| InternVL3-8B | OpenGVLab/InternVL3-8B |
| InternVL3-14B | OpenGVLab/InternVL3-14B |
| InternVL3-38B | OpenGVLab/InternVL3-38B |
| InternVL3-78B | OpenGVLab/InternVL3-78B |
| SmolVLM2-2.2B-Instruct | HuggingFaceTB/SmolVLM2-2.2B-Instruct |
| LLaVA-Video-7B-Qwen2 | lmms-lab/LLaVA-Video-7B-Qwen2 |
| Video-ChatGPT-7B | MBZUAI/Video-ChatGPT-7B |
| LLaVA-OneVision-Qwen2-7B | llava-hf/llava-onevision-qwen2-7b-ov-hf |
| GPT-4o(for modification) | gpt-4o-2024-08-06 |
| GPT-4o(for evaluation) | gpt-4o-2024-11-20 |
| Gemini2.5-flash | gemini-2.5-flash |
| Qwen2-VL-vanilla-rope | Wiselnn/Qwen2-VL-vanilla_rope-128frames-8k-context-330k-llava-video |
| Qwen2-VL-tad-rope | Wiselnn/Qwen2-VL-tad_rope-128frames-8k-context-330k-llava-video |
| Qwen2-VL-m-rope | Wiselnn/Qwen2-VL-m_rope-128frames-8k-context-330k-llava-video |
| Qwen2-VL-video-rope | Wiselnn/Qwen2-VL-videorope-128frames-8k-context-330k-llava-video |

Table 1: Summary of evaluated models and their corresponding official HuggingFace checkpoints.

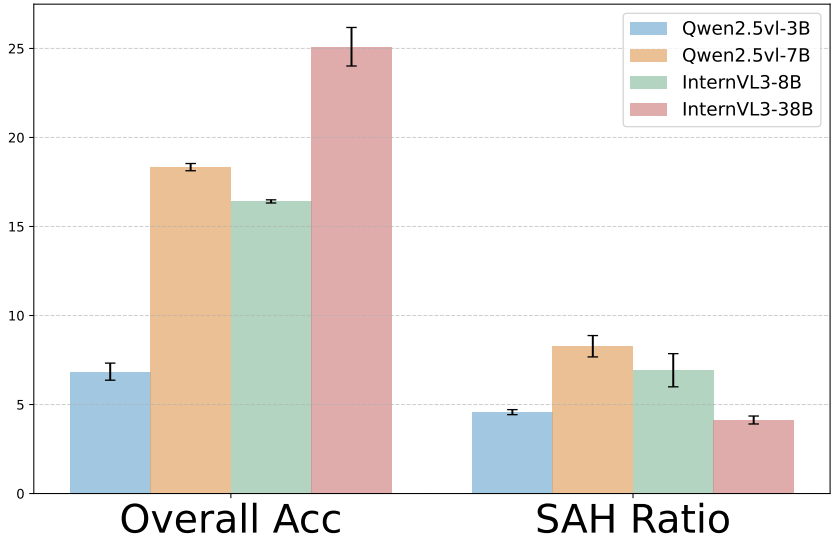

Figure 1: Stability evaluation of ELV-Halluc through repeated experiments. We report overall accuracy and SAH ratio across 3 runs for four representative models of different sizes and series, demonstrating that ELV-Halluc maintains consistent performance.

# More ELV-Halluc Cases

### ELV-Halluc QA Examples

In the following section, we provide detailed examples of question-answer triplets from ELV-Halluc, covering four aspects: visual details, actions, objects, and declarative content. Each triplet consists of three questions: ground truth, in-video hallucination, and out-video hallucination, respectively. We use green, orange, and red to highlight the GT, in-video, and out-video parts, respectively, and enclose them with boxes of the same colors in the video.

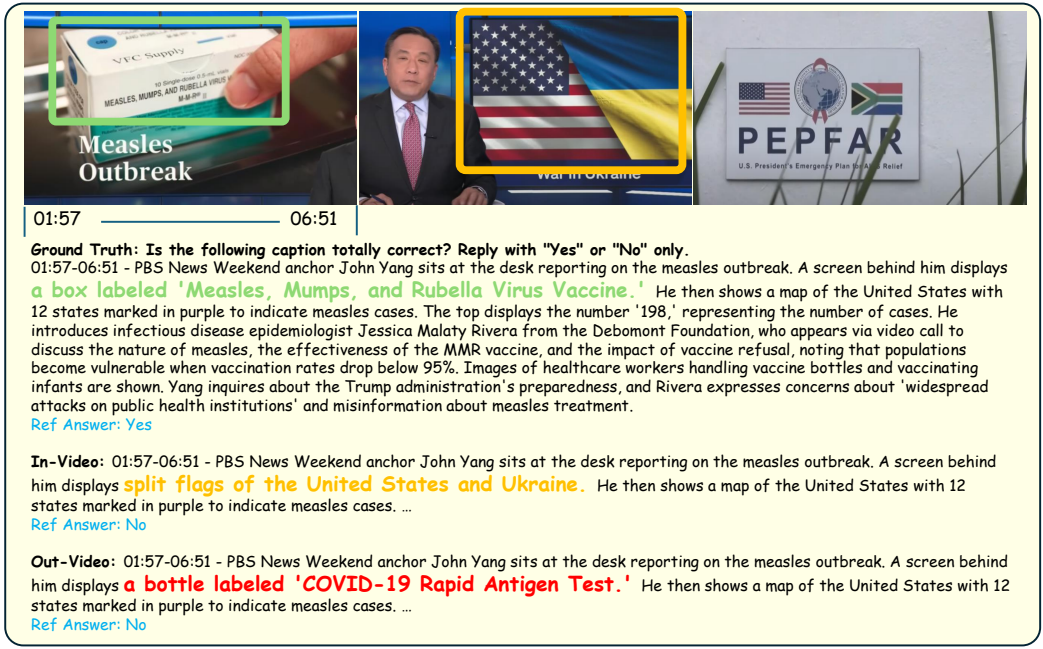

Figure 2: Example of *visual details* QA triplets from ELV-Halluc.

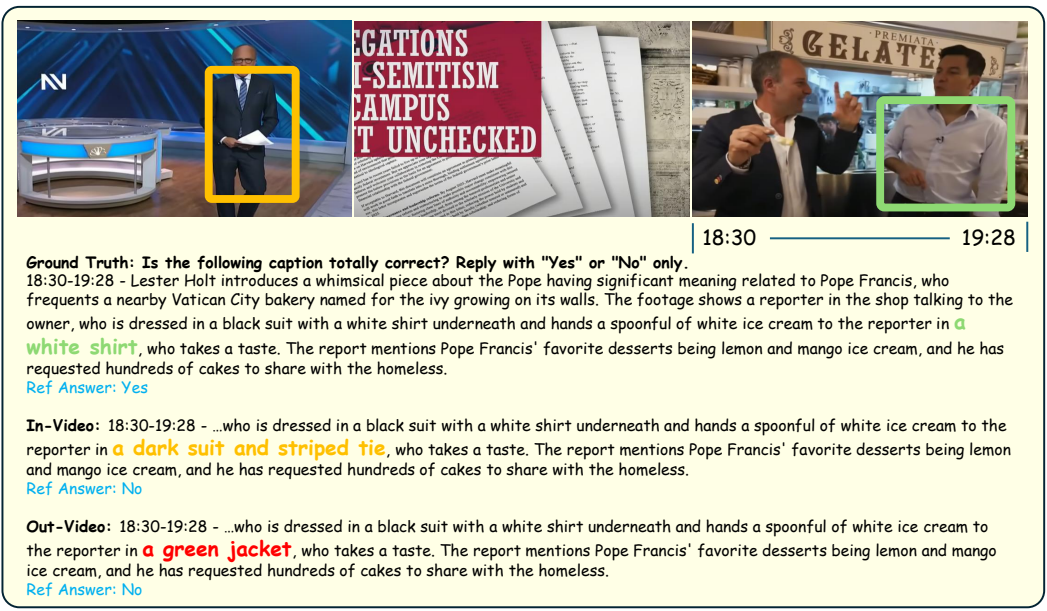

Figure 3: Example of *object* QA triplets from ELV-Halluc.

Figure 4: Example of *action* QA triplets from ELV-Halluc.

Figure 5: Example of *declarative content* QA triplets from ELV-Halluc.

# ELV-Halluc SAH Cases

In the following section, we present specific cases illustrating how SAH happens within ELV-Halluc. The most typical SAH scenario occurs when the ground truth answer is yes, the In-Video response is yes, but the Ou-tVideo response is no. This pattern demonstrates that while the model correctly identifies the hallucination error in the OutVideo scenario, it is misled by the in-video semantic information. These examples clearly show that ELV-Halluc effectively captures and evaluates the presence of SAH. Incorrect answers from the models are highlighted in red.

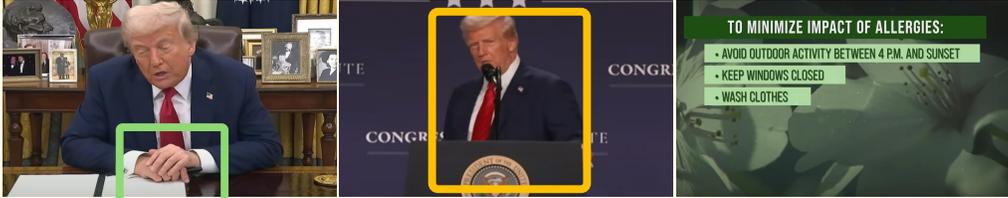

Figure 6: SAH case in ELV-Halluc.

Figure 7: SAH case in ELV-Halluc.

**Ground Truth: Is the following caption totally correct? Reply with "Yes" or "No" only.**
02:37-02:59 - River Plate's No. 29 player Marcos Acuña, wearing a white jersey, takes a corner kick from the left side of the field. The ball curves into the crowded penalty area, and River Plate's No. 8 player Maxi Meza leaps and powerfully heads the ball into the top left corner of the goal, over the yellow goalkeeper's reach. ...
Ref Answer: Yes
Qwen2.5-VL-3B:Yes Qwen2.5-VL-7B :Yes Qwen2.5-VL-72B :Yes InternVL3-14B:Yes InternVL3-78B:Yes

**In-Video:**
02:37-02:59 - River Plate's No. 29 player Marcos Acuña, wearing a white jersey, takes a corner kick from the left side of the field. The ball curves into the crowded penalty area, and River Plate's No. 8 player Driussi ...
Ref Answer: No
Qwen2.5-VL-3B:**Yes** Qwen2.5-VL-7B :**Yes** Qwen2.5-VL-72B :**Yes** InternVL3-14B:No InternVL3-78B:**Yes**

**Out-Video:**
02:37-02:59 - River Plate's No. 29 player Marcos Acuña, wearing a white jersey, takes a corner kick from the left side of the field. The ball curves into the crowded penalty area, and River Plate's No. 8 player Gabriel Batistuta...
Ref Answer: No
Qwen2.5-VL-3B:**Yes** Qwen2.5-VL-7B :**Yes** Qwen2.5-VL-72B :No InternVL3-14B:**Yes** InternVL3-78B:No

Figure 7: SAH case in ELV-Halluc.

**Ground Truth: Is the following caption totally correct? Reply with "Yes" or "No" only.**
00:00-02:19 - A man introduces the imaging of spherical lenses, holding a pencil to show convex and concave lenses. He explains that he will be the object and demonstrates how to accurately draw the principal axis, optical center (O), and focal points (F1, 2F1, F2, 2F2) of a convex lens using a ruler, emphasizing drawing thin lenses.
Ref Answer: Yes
Qwen2.5-VL-3B:Yes Qwen2.5-VL-7B :**No** Qwen2.5-VL-72B :Yes InternVL3-14B:Yes InternVL3-78B:**No**

**In-Video:**
00:00-02:19 - A man introduces the imaging of spherical lenses, holding a pencil to trace rays from the top of his head onto convex and concave lenses. He explains ...
Ref Answer: No
Qwen2.5-VL-3B:**Yes** Qwen2.5-VL-7B :No Qwen2.5-VL-72B :**Yes** InternVL3-14B:**Yes** InternVL3-78B:**Yes**

**Out-Video:**
00:00-02:19 - A man introduces the imaging of spherical lenses, holding a pencil to clean the lenses thoroughly before drawing convex and concave lenses. He explains ...
Ref Answer: No
Qwen2.5-VL-3B:No Qwen2.5-VL-7B :**Yes** Qwen2.5-VL-72B :No InternVL3-14B:No InternVL3-78B:No

Figure 8: SAH case in ELV-Halluc.

**Ground Truth: Is the following caption totally correct? Reply with "Yes" or "No" only.**
01:46-02:19 - On the blue tennis court, Novak Djokovic, wearing a shirt with blue and white patterns, rallied with **Stefanos Tsitsipas**, who was dressed in a white shirt and colorful shorts. **Tsitsipas** attempted a drop shot, but Djokovic quickly rushed forward, extended himself, and hit a sharp-angled backhand winner that landed just inside the line, preventing Tsitsipas from converting a set point and bringing the match to deuce.
Ref Answer: Yes
Qwen2.5-VL-3B:Yes  Qwen2.5-VL-7B :Yes  Qwen2.5-VL-72B :**No**  InternVL3-14B:Yes  InternVL3-78B:Yes

**In-Video:**
01:46-02:19 - On the blue tennis court, Novak Djokovic, wearing a shirt with blue and white patterns, rallied with **Roberto Carballes Baena**, who was dressed in a white shirt and colorful shorts. **Carballes Baena** attempted a drop shot, but...Ref Answer: No
Qwen2.5-VL-3B:**Yes**  Qwen2.5-VL-7B :**Yes**  Qwen2.5-VL-72B :**Yes**  InternVL3-14B:**Yes**  InternVL3-78B:**Yes**

**Out-Video:**
01:46-02:19 - On the blue tennis court, Novak Djokovic, wearing a shirt with blue and white patterns, rallied with **Rafael Nadal**, who was dressed in a white shirt and colorful shorts. **Nadal** attempted a drop shot, but ...
Ref Answer: No
Qwen2.5-VL-3B:No  Qwen2.5-VL-7B :**Yes**  Qwen2.5-VL-72B :No  InternVL3-14B:No  InternVL3-78B:No

Figure 9: SAH case in ELV-Halluc.

# Data Construction Details

In this section, we provide more details on the data construction pipeline, including the prompts used.

As shown in Figure 10, we present the overall word cloud of ELV-Halluc captions to illustrate the diversity of the captions.

Figure 10: Word cloud of all captions in ELV-Halluc.

First, we use the following prompt to instruct Gemini-2.5-Flash to generate scratch captions.

Listing 1: Prompt for gemini-2.5-flash to generate scratch captions.

```
"""You are an expert in understanding videos.

### Task:
You will be provided with a {} video. Your task consists of three main steps:
1. Comprehensively understand the video using both audio and visual information.
2. Segment the video into several parts, while each part contains a isolate and
    complete **{}**.
3. Generate detailed captions for each part according to the following guidelines.

Note: These parts are parallel, which is **a list of distinct but equally
    important elements**.
(e.g., several complete news articles, multiple goals, different cooking steps,
    different movie or TV shows, different parts in vlog, or isolated topics)
```

```
### Guidelines for Video Caption Generation:
Each caption **must include the following aspects**:

1. **Time Range** - Provide the precise time interval of the segment (e.g., "02:15
    0 2 :42").

2. **Visual Details**    Describe specific visual elements in the segment with
    fine granularity:
- Objects: Include object shape, color, texture, written text (OCR), logos, jersey
    numbers, etc.
- Scene Layout: Describe background elements.
- Spatial Relationships: Explain relative positions.

3. **Subject**    Clearly identify the main subject(s) in the segment.

4. **Action**    Describe physical activities with verbs.

5. **Event**    Describe what happens as a complete narrative.

### Output Instructions:
1. Captions should be coherent and complete sentences.
2. Ignore redundant and uninformative content such as static frames, repetitive
    scenes, intros/outros, idle moments, and summary frames at the beginning or
    end of the video.
3. Ignore transitional or filler content between each major segment.
4. Each video must be conclude into less than 10 parts.

### Output Format:
You must strictly follow the output format below and output nothing beyond the
    specified structure.
```json
    {{
        "part_1": "mm:ss-mm:ss - ...",
        "part_2": "mm:ss-mm:ss - ...",
        ...
    }}
```
"""
```

After manual checking, we prompt GPT-4o to inject in-video and out-of-video hallucinations related to four aspects into the ground truth captions. The prompts are shown below.

Listing 2: Prompt for GPT-4o to inject visual detail hallucinations.

```
You are a professional caption generation assistant.
You will be provided with captions for different parts of the same video. Your
    task is to generate hallucinated captions following the instructions below.

### Current Part Caption (Original Caption):
{current_part}

### Other Parts Captions:
{other_parts}

## Instructions:
1. Break down the current part caption into the following components:
    - **Visual details**: attributes such as color, shape, size, patterns, spatial
        relationships, or on-screen text (OCR).
    - **Objects**: refers to humans or physical objects involved in the caption.
    - **Actions**: refers to the key activity or motion being performed.
    - **Declarative Content**: refers to high-level descriptive or propositional
        statements that summarize a situation, assert an outcome, or convey a
        belief, stance, or result. These are not concrete actions or events.
    - **Time range**: the timestamp segment covered by this caption.
```

9

```
2. Identify a **visual detail** in the current part caption.

3. Replace that visual detail using:
   - **in_video hallucination**: Replace it with a visual detail that appears in *
     other parts captions* of the same video.
   - **out_video hallucination**: Replace it with a visual detail that *never
     appears in any part captions* of this video.

4. The hallucinated captions must remain **plausible and reasonable**, so that a
    model cannot reliably judge which is correct (among original, in_video, and
    out_video) without actually seeing the video.

### IMPORTANT:
- You must strictly preserve all the rest of the current part caption (including
    its time range and sentence structure). Only the selected **visual detail** is
     to be replaced.
- The replacement must be logically consistent and naturally integrated, with no
    contradictions or cues that obviously reveal the modification.
- The replacement must change the semantics of the original caption, and cannot
    use synonyms.

### Output Format:
```json
{{
  "in_video": "in_video caption here",
  "out_video": "out_video caption here"
}}
```
```

Listing 3: Prompt for GPT-4o to inject object hallucinations.

```
You are a professional caption generation assistant.
You will be provided with captions for different parts of the same video. Your
    task is to generate hallucinated captions following the instructions below.

### Current Part Caption (Original Caption):
{current_part}

### Other Parts Captions:
{other_parts}

## Instructions:
1. Break down the current part caption into the following components:
   - **Visual details**: attributes such as color, shape, size, patterns, spatial
       relationships, or on-screen text (OCR).
   - **Objects**: refers to humans or physical objects involved in the caption.
   - **Actions**: refers to the key activity or motion being performed.
   - **Declarative Content**: refers to high-level descriptive or propositional
       statements that summarize a situation, assert an outcome, or convey a
       belief, stance, or result. These are not concrete actions or events.
   - **Time range**: the timestamp segment covered by this caption.

2. Identify a **object** in the current part caption.

3. Replace that object using:
   - **in_video hallucination**: Replace it with an object that appears in *other
       parts captions* of the same video.
   - **out_video hallucination**: Replace it with an object that *never appears in
        any part captions* of this video.

4. The hallucinated captions must remain **plausible and reasonable**, so that a
    model cannot reliably judge which is correct (among original, in_video, and
```

```
        out_video) without actually seeing the video.

### IMPORTANT:
- You must strictly preserve all the rest of the current part caption (including
    its time range and sentence structure). Only the selected **object** is to be
    replaced.
- The replacement must be logically consistent and naturally integrated, with no
    contradictions or cues that obviously reveal the modification.
- The replacement must change the semantics of the original caption, and cannot
    use synonyms.

### Output Format:
```json
{{
  "in_video": "in_video caption here",
  "out_video": "out_video caption here"
}}
```
```

Listing 4: Prompt for GPT-4o to inject action hallucinations.

```
You are a professional caption generation assistant.
You will be provided with captions for different parts of the same video. Your
    task is to generate hallucinated captions following the instructions below.

### Current Part Caption (Original Caption):
{current_part}

### Other Parts Captions:
{other_parts}

## Instructions:
1. Break down the current part caption into the following components:
    - **Visual details**: attributes such as color, shape, size, patterns, spatial
        relationships, or on-screen text (OCR).
    - **Objects**: refers to humans or physical objects involved in the caption.
    - **Actions**: refers to the key activity or motion being performed.
    - **Declarative Content**: refers to high-level descriptive or propositional
        statements that summarize a situation, assert an outcome, or convey a
        belief, stance, or result. These are not concrete actions or events.
    - **Time range**: the timestamp segment covered by this caption.

2. Identify an **action** in the current part caption.

3. Replace that action using:
    - **in_video hallucination**: Replace it with an action that appears in *other
        parts captions* of the same video.
    - **out_video hallucination**: Replace it with an action that *never appears in
        any part captions* of this video.

4. The hallucinated captions must remain **plausible and reasonable**, so that a
    model cannot reliably judge which is correct (among original, in_video, and
    out_video) without actually seeing the video.

### IMPORTANT:
- You must strictly preserve all the rest of the current part caption (including
    its time range and sentence structure). Only the selected **action** is to be
    replaced.
- The replacement must be logically consistent and naturally integrated, with no
    contradictions or cues that obviously reveal the modification.
- The replacement must change the semantics of the original caption, and cannot
    use synonyms.
```

11

```
### Output Format:
```json
{{
  "in_video": "in_video caption here",
  "out_video": "out_video caption here"
}}
```
```

Listing 5: Prompt for GPT-4o to inject declarative content hallucinations.

```
You are a professional caption generation assistant.
You will be provided with captions for different parts of the same video. Your
    task is to generate hallucinated captions following the instructions below.

### Current Part Caption (Original Caption):
{current_part}

### Other Parts Captions:
{other_parts}

## Instructions:
1. Break down the current part caption into the following components:
    - **Visual details**: attributes such as color, shape, size, patterns, spatial
        relationships, or on-screen text (OCR).
    - **Objects**: refers to humans or physical objects involved in the caption.
    - **Actions**: refers to the key activity or motion being performed.
    - **Declarative Content**: refers to high-level descriptive or propositional
        statements that summarize a situation, assert an outcome, or convey a
        belief, stance, or result. These are not concrete actions or events.
    - **Time range**: the timestamp segment covered by this caption.

2. Identify a **declarative content** in the current part caption.

3. Replace that declarative content using:
    - **in_video hallucination**: Replace it with a declarative content that
        appears in *other parts captions* of the same video.
    - **out_video hallucination**: Replace it with a declarative content that *
        never appears in any part captions* of this video.

4. The hallucinated captions must remain **plausible and reasonable**, so that a
    model cannot reliably judge which is correct (among original, in_video, and
    out_video) without actually seeing the video.

### IMPORTANT:
- You must strictly preserve all the rest of the current part caption (including
    its time range and sentence structure). Only the selected **declarative
    content** is to be replaced.
- The replacement must be logically consistent and naturally integrated, with no
    contradictions or cues that obviously reveal the modification.
- The replacement must change the semantics of the original caption, and cannot
    use synonyms.

### Output Format:
```json
{{
  "in_video": "in_video caption here",
  "out_video": "out_video caption here"
}}
```
```

Finally, we prompt GPT-4o to recheck all modified captions, and we retain only the question pairs that are verified as true.

Listing 6: Prompt for GPT-4o to check hallucinated captions.

```
You are a strict video caption checker.

You will be given multiple captions from different parts in a same video.
Each part has a ground-truth caption and two hallucinated captions: `in_video` (
    containing a modification that should come from another part of the same video
    ) and `out_video` (containing a plausible but fabricated modification that
    does not appear in any part of the video).
The hallucination aspect of this part is **{aspect}**.

First, identify what information has been added or changed in the hallucinated
    captions compared to the ground_truth.
Second, verify whether:
1. The **in_video** caption introduces a/an **{aspect}** modification that **
    semanticly appears** in at least one **other part's** `ground_truth` (
    excluding the current part).
2. The **out_video** caption introduces a/an **{aspect}** modification that **
    semanticly does NOT appear** in any of the ground_truth captions.
Focus on the differences between the hallucinated captions and the original
    ground_truth.
Please return your judgment strictly follows the below JSON format, don't provide
    any reasons, explanations, verifications:
```json
{{
  "in_video_valid": true/false,
  "out_video_valid": true/false
}}
```
Ground Truth:
{ground_truth}

In-Video Hallucination:
{in_video}

Out-Video Hallucination:
{out_video}

Other Parts' Ground Truths:
{other_parts_gt}
```



\end{document}